\documentclass{article} % For LaTeX2e
\usepackage{iclr2017_conference,times}
\usepackage[colorlinks=true,linkcolor=blue,citecolor=blue]{hyperref}
\usepackage{amsmath,amssymb,array,algorithm}
\usepackage[algo2e]{algorithm2e}
\usepackage{booktabs,color,caption,graphicx,subcaption}

\title{SoftTarget Regularization\\ \\ An effective technique to reduce over-fitting in Neural Networks}

\author{Armen Aghajanyan \\
Dimensional Mechanics\\
Bellevue, WA 98007, USA \\
\texttt{armen.aghajanyan@dimensionalmechanics.com}}

\begin{document}

\maketitle

\begin{abstract}
	Deep neural networks are learning models with a very high capacity and
	therefore prone to over-fitting. Many regularization techniques such as
	Dropout, DropConnect, and weight decay all attempt to solve the problem of
	over-fitting by reducing the capacity of their respective models
	\citep{Srivastava2014}, \citep{Wan2013}, \citep{Krogh1992}. In this paper we
	introduce a new form of regularization that guides the learning problem in a
	way that reduces over-fitting without sacrificing the capacity of the model.
	The mistakes that models make in early stages of training carry information
	about the learning problem. By adjusting the labels of the current epoch of
	training through a weighted average of the real labels, and an exponential
	average of the past soft-targets we achieved a regularization scheme as
	powerful as Dropout without necessarily reducing the capacity of the model,
	and simplified the complexity of the learning problem. SoftTarget
	regularization proved to be an effective tool in various neural network
	architectures.
\end{abstract}

\section{Introduction}
Many regularization techniques have been created to rectify the problem of
over-fitting in deep neural networks, but the majority of these methods reduce
models capacities to force them to learn general enough features. For example,
Dropout reduces the amount of learn-able parameters by randomly dropping
activations, and DropConnect extends this idea by randomly dropping weights
\citep{Srivastava2014}, \citep{Wan2013}. Weight decay regularization reduces the
capacity of the model, not by dropping learn-able parameters, but by reducing
the space of viable solutions \citep{Krogh1992}.

\subsection{Motivation}
Hinton has shown that soft-labels, or labels predicted from a model contain more
information that binary hard labels due to the fact that they encode similarity
measures between the classes \citep{Hinton2015}. Incorrect labels tagged by the
model describe co-label similarities, and these similarities should be evident
in future stages of learning, even if the effect is diminished. For example,
imagine training a deep neural net on a classification dataset of various dog
breeds. In the initial few stages of learning the model will not accurately
distinguish between similar dog-breeds such as a Belgian Shepherd versus a
German Shepherd. This same effect, although not so exaggerated, should appear
in later stages of training. If, given an image of a German Shepherd, the model
predicts the class German Shepherd with a high-accuracy, the next highest
predicted dog should still be a Belgian Shepherd, or a similar looking dog.
Over-fitting starts to occur when the majority of these co-label effects begin
to disappear. By forcing the model to contain these effects in the
later stages of training, we reduced the amount of over-fitting.

\subsection{Method}
Consider the standard supervised learning problem. Given a dataset containing
inputs and outputs, $X$ and $Y$, a regularization function $R$ and a model
prediction function $F$ we attempted to minimize the loss function
$\mathcal{L}$ given by:

\begin{equation}
	\mathcal{L}(X,Y) = \frac{1}{N} \sum_{i=0}^{N}
	\mathcal{L}_i(\mathcal{F}(X_i, \mathbf{W}),Y_i) + \lambda R(\mathbf{W})
\end{equation}

where $\mathbf{W}$ are the weights in $\mathcal{F}$ that are adjusted to
minimize the loss function, and $\lambda$ controls the effect of the
regularization function. For our method to fit into the supervised learning
scheme we altered the optimization problem by adding a time dimension $(t)$ to
the loss function:

\begin{equation}
	\mathcal{L}^{t}(X,Y) = \frac{1}{N} \sum_{i=0}^{N}
	\mathcal{L}_i^{t}(\mathcal{F}(X_i, \mathbf{W}),Y_i) + \lambda R(\mathbf{W})
\end{equation}

SoftTarget regularization requires into two steps: first, we kept an
exponential moving average of past labels $\hat{Y}^t$, and second, we updated
the current epochs label $Y^{t}_{c}$ through a weighted average of the
exponential moving average of past labels and of the true hard labels:

\begin{align}
	\hat{Y}^t & = \beta \hat{Y}^{t-1} + (1-\beta)\mathcal{F}(X_i, \mathbf{W}) \\
	Y_c^{t}   & = \gamma \hat{Y}^t + (1-\gamma)Y
\end{align}

Here, $\gamma$ and $\beta$ are hyper-parameters that can be tuned to specific
applications. The loss function then becomes:

\begin{equation}
	\mathcal{L}^{t}(X,Y) = \frac{1}{N} \sum_{i=0}^{N}
	\mathcal{L}_i^{t}(\mathcal{F}(X_i, \mathbf{W}), Y_c^{t}) + \lambda
	R(\mathbf{W})
\end{equation}

The algorithm also contains a `burn-in' period, where no SoftTarget
regularization is done and the model is trained freely in order to learn the
basic co-label similarities. We will denote the number of epochs trained freely
as $n_b$, and the total number of epochs as $n$. Experimentally we also
discovered that it is sometimes best to run the network for more than one epoch
on a single $Y_c$, so we will denote $n_t$ as the number of epochs per every
time-step. We have provided the pseudo-code in Algorithm~\ref{algo1}.

\begin{algorithm}[H]
	\SetKwInOut{Input}{input}
	\Input{$X, Y, \mathcal{F}, \mathcal{G}, \beta, \gamma, n_b, n_t, n$}
	\BlankLine{}
	\BlankLine{}

	$\mathcal{F} \leftarrow \mathcal{G}(\mathcal{F}, \{X, Y\}, n_b)$\;
	\BlankLine{}
	$\hat{Y}^0 \leftarrow \mathcal{F}(X_i, \mathbf{W})$, $t \leftarrow 1$
	\;{}
	\BlankLine{}
	\For{$i\leftarrow 0$ \KwTo{} $\frac{n - n_b}{n_t}$}{
		\begin{math}
			\hat{Y}^t = \beta \hat{Y}^{t-1} + (1-\beta)\mathcal{F}(X_i, \mathbf{W})\\
			Y_c^{t} = \gamma \hat{Y}^t + (1-\gamma)Y \\
		\end{math}

		$\mathcal{F} \leftarrow \mathcal{G}(\mathcal{F}, \{X, Y_c^{t}\},
		n_t)$, $t \leftarrow t + 1$ \;
	}
	\caption{SoftTarget Regularization\label{algo1}}
\end{algorithm}

Here $\mathcal{G}$ represents the training of the neural network, taking in a
model $\mathcal{F}$, dataset $\{X, Y\}$ and an integer representing number of
epochs.

A large $n_t$ allows the network to learn a better mapping to the intermediate
soft-labels and therefore allows the regularization to be more effective. But
increasing $n_t$ has a diminishing effect, because as $n_t$ becomes large the
network begins to over-fit to those soft-labels, and reduces the effect of the
regularization, as well as increasing the training time of the network
significantly. $n_t$ should be optimized experimentally through standard
hyper-parameter optimization practices. We found $n_t = \{1,2\}$ to work best through standard grid hyper-parameter optimization \citep{Bergstra:2012:RSH:2503308.2188395}.
The small $n_t$ insures that the model does not overfit to the intermediate representation introduced by SoftTarget.

Through hyper-parameter optimization the same range of $\{1,2\}$ was found to be optimal in the experiments we ran for $n_b$. A small $n_b$ insures that the
co-label similarities captured by SoftTarget would not have been affected by any type of overfitting. This insures that as the experiments are further ran the true co-label similarties
are propagated correctly. More complicated learning scenarias where the amount of labels and data is greater, the chances of corruption in co-label similarties is reduced and therefore larger
$n_b$ can be choosen.

\subsection{Similarities to Other Methods}
Other methods similar to this are specific to the case where the $\beta$
hyper-parameter is set to zero, with no burn-in period.

\begin{itemize}
	\item Reed et al.\ study the specific case of the SoftTarget method described
	      above with the $\beta$ parameter set to zero \citep{Reed2014}. They
	      focus on the capability of the network to be robust to noise, rather
	      than the regularization abilities of the method.

	\item Grandvalet and Bengio have proposed minimum entropy regularization in
	      the setting of semi-supervised learning \citep{Grandvalet2005}. This
	      algorithm changes the categorical cross-entropy loss to force the
	      network to make predictions with high degrees of confidence on the
	      unlabeled portion of the dataset. Assuming cross-entropy loss with
	      SoftTarget normalization with a zero burn-in period, and zero $\beta$,
	      our algorithm becomes equivalent to a softmax regression with minimum
	      entropy regularization.

	\item Another similar approach to minimum entropy regularization is an
	      approach called pseudo-labeling. Pseudo-labeling tags unlabeled data
	      with the class predicted highest by a learning model
	      \citep{Lee2013}. No soft-targets are kept, instead the
	      predicted label is binarized, i.e.\ the highest class is labeled with a
	      value of one, and every other class is labeled with a value of zero.
	      These hard pseudo-labels are then fed as input to the model.

	\item Hinton et al described the power of soft targets in the use of transferring knowledge from one model to another, usually to a model that contains less parameters \citep{Hinton2015}. Soft-Target regularization can be interpreted as weighted distillation where the donor model is the state of the model at some previous time in training, and the weighting target are the hard-targets.
\end{itemize}

\section{Experiments}
We conducted experiments in python using the Theano and Keras libraries
\citep{TheTheanoDevelopmentTeam2016}, \citep{chollet2015}. All of our code ran
on a single Nvidia Titan X GPU, while using the CnMEM and cuDNN (5.103)
extensions, and we visualized our results using matplotlib \citep{Hunter2007}.
We used the same seed in all our calculations to insure the starting weights
were equivalent in every set of experiments. The only source of randomness
stemmed from the non-deterministic behavior of the cuDNN libraries.

\subsection{MNIST}
We first considered the famous MNIST dataset \citep{LeCun1998}. For each of the
experiments discussed below, we performed a random grid-search over the
hyper-parameters of the optimization algorithm, and a very small brute force
grid search was done for the hyper-parameters of SoftTarget regularization. We compared our results to the cases where the
hyper-parameters resulted in the best performance of the vanilla neural network
without SoftTarget regularization. All of our reported values were computed on
the standardized test portion of the MNIST dataset, as provided by the Keras
library. The networks were trained strictly on the training portion of the
dataset. We tested on eight different architectures, with four combinations of
every architecture. The four combinations stem from testing each architecture
via a combination of: no regularization, Dropout, SoftTarget, and
Dropout+SoftTarget regularization.

We used a fully connected network, with a varying amount of hidden layers, and a
set constant of neurons throughout each layer. Dropout was not introduced at
the input layer, but was introduced at every layer after that. All of the
layers activations we're rectified linear units (ReLu), except for the final layer which was a SoftMax. The net was trained using a categorical cross-entropy loss, and the ADADELTA optimization method. \citep{Zeiler2012}.

The frozen hyper-parameters for the SoftTarget regularization were: $n_b =
2,n_t=2,n=100, \beta = 0.7, \gamma = 0.5$. Our results are described in
Table~\ref{tab:mnistcomp}. We described the nets using the notation: 4
$\leftarrow$ 256 denoting a 4 hidden layer neural network, with each of the
hidden layers having 256 units. We reported the minimum loss during training, the loss at the 100th epoch, and the maximum accuracy reached respectively.

\begin{table}[htb]
	\caption{\label{tab:mnistcomp} MNIST Comparison: minimum loss, loss at 100th epoch, max accuracy}
	\resizebox{\textwidth}{!}
	{%
		\begin{tabular}{ccccccc}
			\toprule
			Net                 & Vanilla     & SoftTarget                                       & SoftTarget+Dropout (0.2)                         & SoftTarget+Dropout (0.5) & Dropout (0.2) & Dropout (0.5) \\ \toprule
			4 $\leftarrow$ 256  & $0.076|0.208|0.981$ & $\textcolor{green}{0.063}|\textcolor{blue}{0.095}|0.982$ & $0.068|0.102|\textcolor{purple}{0.989}$              & $0.114|0.143|0.974$      & $0.081|0.150|0.983$   & $0.137|0.198|0.978$   \\ \midrule
			5 $\leftarrow$ 512  & $0.077|0.206|0.984$ & $\textcolor{green}{0.056}|\textcolor{blue}{0.069}|\textcolor{purple}{0.986}$ & $0.060|0.113|0.985$              & $0.101|0.117|0.978$      & $0.087|0.164|0.984$   & $0.088|0.170|0.976$   \\ \midrule
			6 $\leftarrow$ 256  & $0.199|0.334|0.979$ & $\textcolor{green}{0.063}|\textcolor{blue}{0.092}|\textcolor{purple}{0.990}$ & $0.075|0.101|0.982$              & $0.148|0.150|0.985$      & $0.101|0.202|0.981$   & $0.086|0.252|0.970$   \\ \midrule
			6 $\leftarrow$ 512  & $0.079|0.241|0.981$ & $\textcolor{green}{0.056}|\textcolor{blue}{0.068}|\textcolor{purple}{0.990}$ & $0.064|0.131|0.985$              & $0.131|0.159|0.977$      & $0.089|0.190|0.981$   & $0.152|0.339|0.978$   \\ \midrule
			7 $\leftarrow$ 256  & $0.092|0.246|0.981$ & $\textcolor{green}{0.065}|\textcolor{blue}{0.079}|\textcolor{purple}{0.985}$ & $0.083|0.100|0.983$              & $0.207|0.222|0.978$      & $0.108|0.215|0.977$   & $0.216|0.232|0.968$   \\ \midrule
			7 $\leftarrow$ 512  & $0.090|0.244|0.982$ & $\textcolor{green}{0.056}|\textcolor{blue}{0.077}|\textcolor{purple}{0.985}$ & $0.071|0.107|\textcolor{purple}{0.985}$              & $0.172|0.211|0.978$      & $0.099|0.236|0.983$   & $0.175|0.383|0.974$   \\ \midrule
			3 $\leftarrow$ 256  & $0.074|0.197|0.981$ & $\textcolor{green}{0.064}|0.105|0.985$   & $0.068|\textcolor{blue}{0.092}|\textcolor{purple}{0.990}$            & $0.109|0.145|0.975$      & $0.079|0.121|0.985$   & $0.118|0.155|0.980$   \\ \midrule
			3 $\leftarrow$ 1024 & $0.065|0.138|0.982$ & $0.055|0.088|0.983$                      & $\textcolor{green}{0.054}|\textcolor{blue}{0.084}|\textcolor{purple}{0.990}$ & $0.072|0.112|0.982$ & $0.065|0.138|0.985$   & $0.088|0.137|0.983$   \\ \midrule
			3 $\leftarrow$ 2048 & $0.065|0.139|0.982$ & $0.053|0.104|0.983$                      & $\textcolor{green}{0.052}|\textcolor{blue}{0.072}|\textcolor{purple}{0.990}$ & $0.060|0.096|0.990$ & $0.071|0.141|0.978$   & $0.088|0.104|0.987$   \\ \bottomrule

		\end{tabular}%
	}
\end{table}

\begin{figure}
	\begin{subfigure}[htb]{200px}
		\centering
		\includegraphics[width=200px]{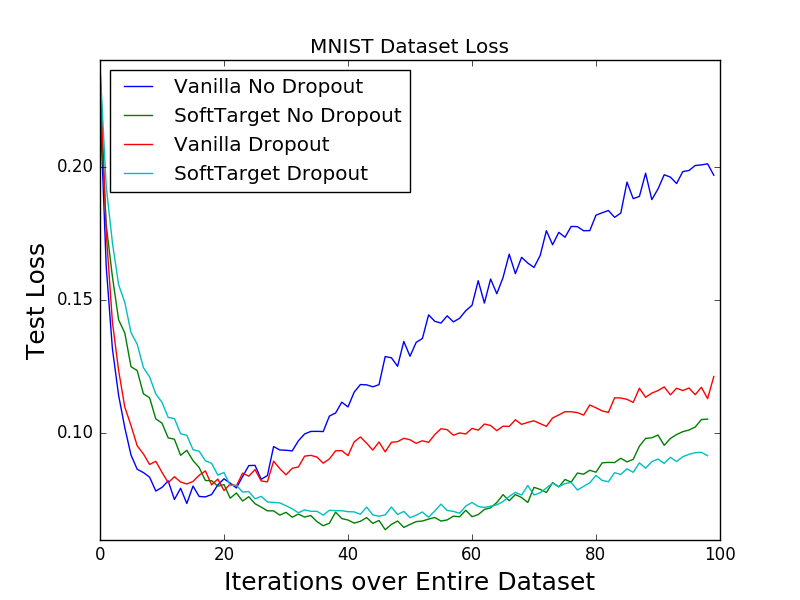}
		\caption{3 Layers, 256 Units, Dropout=0.2}
	\end{subfigure}%
	\begin{subfigure}[htb]{200px}
		\centering
		\includegraphics[width=200px]{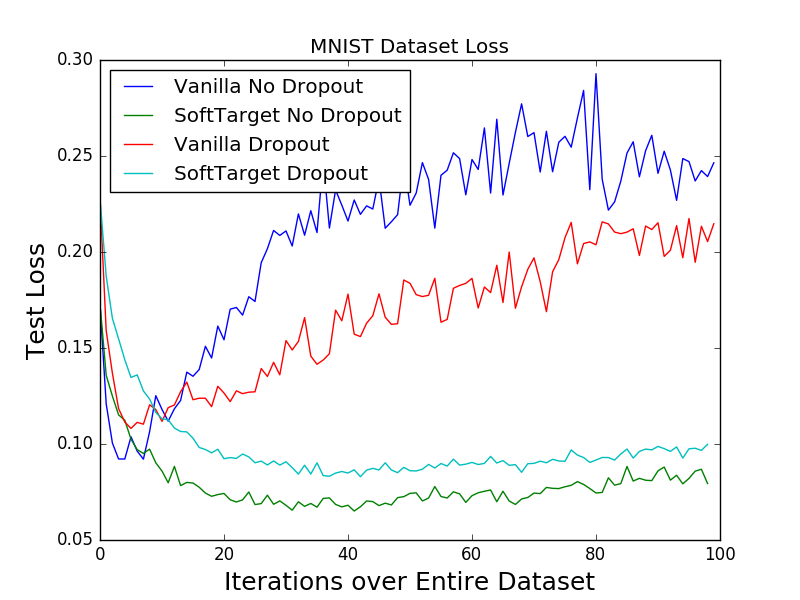}
		\caption{7 Layers, 256 Units, Dropout=0.2}
	\end{subfigure}%
	\caption{\label{fig:reg}Regularization applied to multilayer neural networks.}
\end{figure}

In all our experiments, the best performing regularization for all of the
architectures described above included SoftTarget regularization. Two
representative results are plotted in Figure~\ref{fig:reg} for a shallow (three
layer) and deep (seven layer) neural network. We saw that for deep neural
networks (greater than three layers) SoftTarget regularization outperformed all
the other regularization schemes. For shallow (three layer) neural networks
SoftTarget+Dropout was the optimal scheme.

\subsection{CIFAR-10}
We then considered the CIFAR-10 dataset \citep{Krizhevsky2009}, comparing
various combinations of SoftTarget, Dropout and BatchNormalization (BN)
\citep{Ioffe2015}. BatchNormalization has been shown to have a regularization
effect on neural networks due to the noise inherent to the mini-batch
statistics. We ran each configuration of the network through sixty iterations
through the whole training set. The complete architecture used was:

Input $\rightarrow$ Convolution (64,3,3) $\rightarrow$ BN $\rightarrow$ ReLU
$\rightarrow$ Convolution (64,3,3) $\rightarrow$ BN $\rightarrow$ ReLU
$\rightarrow$ MaxPooling ((3,3), (2,2)) $\rightarrow$ Dropout ($p$)
$\rightarrow$ Convolution (128,3,3) $\rightarrow$ BN $\rightarrow$ ReLU
$\rightarrow$ Convolution (128,3,3) $\rightarrow$ BN $\rightarrow$ ReLU
$\rightarrow$ MaxPooling ((3,3), (2,2)) $\rightarrow$ Dropout ($p$)
$\rightarrow$ Convolution (256,3,3) $\rightarrow$ BN $\rightarrow$ ReLU
$\rightarrow$ Convolution (256,1,1) $\rightarrow$ BN $\rightarrow$ ReLU
$\rightarrow$ Convolution (256,1,1) $\rightarrow$ BN $\rightarrow$ ReLU
$\rightarrow$ Dropout ($p$) $\rightarrow$ AveragePooling ((6,6)) $\rightarrow$
Flatten () $\rightarrow$ Dense (256) $\rightarrow$ BN $\rightarrow$ ReLU
$\rightarrow$ Dense (256) $\rightarrow$ BN $\rightarrow$ ReLU $\rightarrow$
Dense (256) $\rightarrow$ SoftMax.

where: Convolution (64,3,3) signifies the convolution operator with 64 filters,
and a kernel size of 3 by 3, MaxPooling ((3,3), (2,2)) represents the
max-pooling operation with a kernel size of 3 by 3, and a stride of 2 by 2,
AveragePooling ((6,6)) represents the average pooling operator with a kernel
size of 6 by 6, Flatten represents a flattening of the tensor into a matrix,
and Dense (256) a fully-connected layer \citep{Krizhevsky2012}, \citep{Scherer2010}. In our results, when we note that BN or Dropout weren't used, we simply omitted those layers from the architecture. We trained the networks using ADADELTA on the cross-entropy loss, using the same SoftTarget hyper-parameters we reported for the MNIST dataset. Our results are summarized in Table~\ref{tab:cifar}. The first column specifies the amount of Dropout used on the combinations listed in the next columns. As with the MNIST experiments, we reported the minimum loss during training, and the loss at the 100th epoch.

\begin{table}[]
	\centering
	\caption{\label{tab:cifar}CIFAR-10 Comparison}
	\begin{tabular}{ccccc}
		\toprule
		Amount of Dropout & BN          & SoftTarget                                       & Just Dropout                   & SoftTarget+BN                                    \\ \midrule
		0                 & $0.731|1.876|0.821$ & $0.511|0.592|0.838$                      & $0.595|1.120|0.831$            & $\textcolor{green}{0.502}|\textcolor{blue}{0.540}|\textcolor{purple}{0.840}$ \\ \midrule
		0.2               & $0.517|0.855|0.866$ & $0.450|0.501|0.854$                      & $0.518|0.706|0.848$            & $\textcolor{green}{0.408}|\textcolor{blue}{0.410}|\textcolor{purple}{0.872}$ \\ \midrule
		0.4               & $0.452|0.596|0.865$ & $0.434|0.478|0.865$                      & $0.463|0.543|0.851$            & $\textcolor{green}{0.403}|\textcolor{blue}{0.432}|\textcolor{purple}{0.871}$ \\ \midrule
		0.6               & $0.487|0.560|0.855$ & $\textcolor{green}{0.474}|\textcolor{blue}{0.488}|0.845$ & $0.480|0.550|0.835$                    & $0.489|0.526|\textcolor{purple}{0.870}$                      \\ \midrule
		0.8               & $0.677|0.741|0.772$ & $0.672|\textcolor{blue}{0.695}|\textcolor{purple}{0.774}$     & $\textcolor{green}{0.620}|0.714|0.737$ & $0.721|0.777|0.756$                     \\ \bottomrule
	\end{tabular}%
\end{table}

\begin{figure}[t]
	\begin{center}
		\begin{subfigure}[b]{140px}
			\centering
			\includegraphics[width=140px]{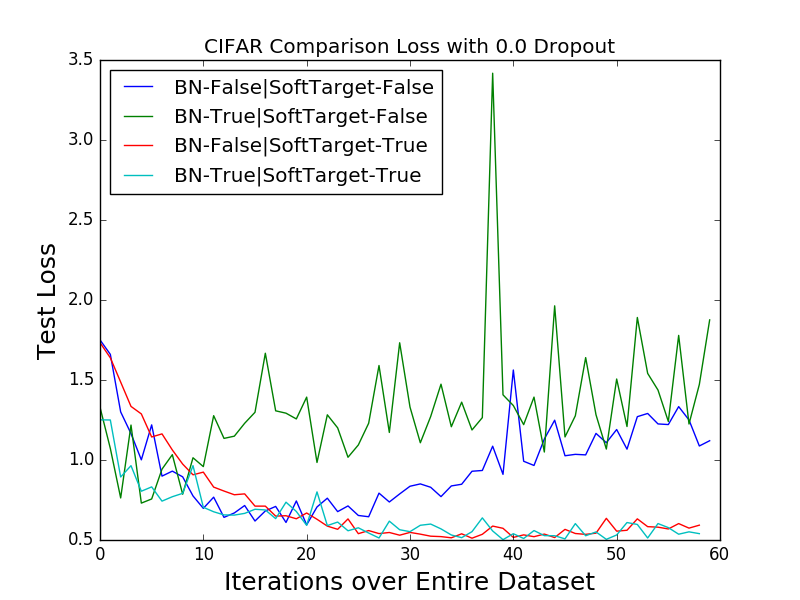}
			\caption{No Dropout}
		\end{subfigure}
		\begin{subfigure}[b]{140px}
			\centering
			\includegraphics[width=140px]{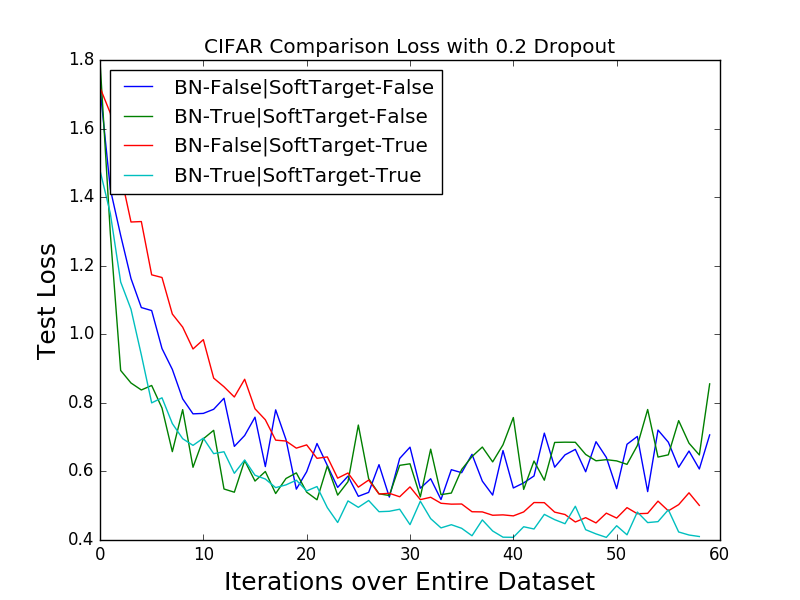}
			\caption{Dropout=0.2}
		\end{subfigure}\\
		\begin{subfigure}[b]{140px}
			\centering
			\includegraphics[width=140px]{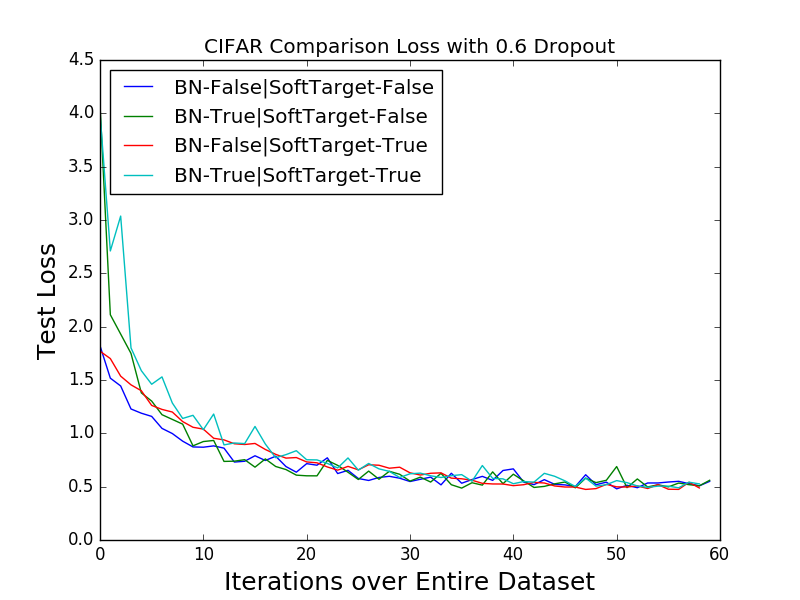}
			\caption{Dropout=0.6}
		\end{subfigure}
		\begin{subfigure}[b]{140px}
			\centering
			\includegraphics[width=140px]{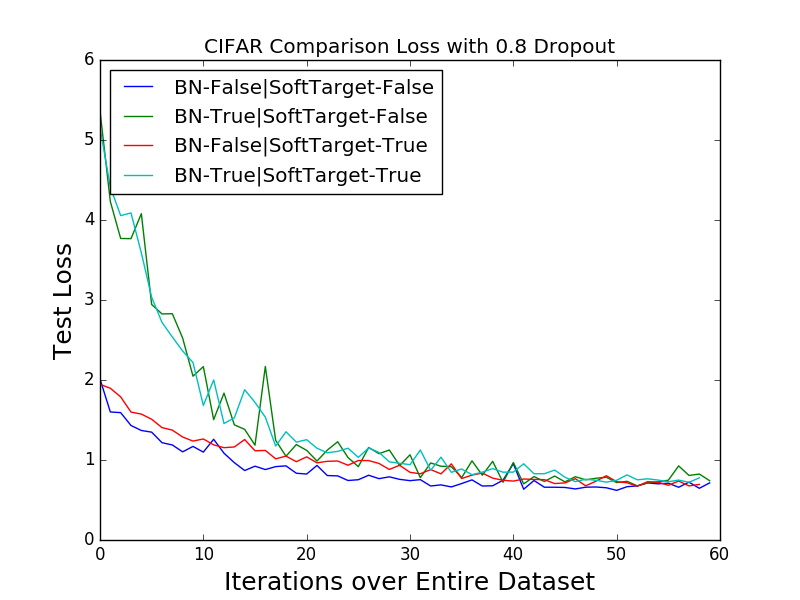}
			\caption{Dropout=0.8}
		\end{subfigure}
	\end{center}
\end{figure}

The use of SoftTarget regularization resulted in the lowest loss in four out of
the five experiments on this architecture, and resulted in the lowest last epoch
loss value and highest accuracy in all five of the experiments. As the dropout rate is increased the need for any other type of regularization is decreased. However, by increasing the rate of dropout, the resulting loss is increased because of the reduced capacity of the network. SoftTarget regularization allowed a lower dropout rate
to be used, and this lowered the test error.

\subsection{SVHN}
Finally, we considered the Street View House Numbers (SVHN) dataset, consisting
of various images mapping to one of ten digits \citep{Netzer2011}. This is
similar to the MNIST dataset, but is much more organic in nature, as these
images contain much more natural noise, such as lighting conditions and camera
orientation. We tested residual networks in four configurations: No
regularization, Batch Normalization (BN), SoftTarget, and BN+SoftTarget
\citep{1512.03385}. Our architecture consisted of the same building blocks as
the residual network outlined by He et al., consisting of identity and
convolution blocks \citep{He2015}. Identity blocks are blocks that do not
contain a convolution layer at the shortcut, while convolution blocks do. In
our notation I (3,[16,16,32], BN) will mean an identity block with an
intermediate square convolution kernel size of 3, with three convolution blocks
of size 16, 16 and 32. The outer convolutions contain kernel sizes of 1. C
(3,[16,16,32], BN) contains the same initial architecture as I (3,[16,16,32])
but an additional convolution layer of size 32 at the shortcut connection. All
of these blocks contained the rectified linear function as their activation,
and BN prior to activation. Our final architecture was:

Input $\rightarrow$ ZeroPadding (3,3) $\rightarrow$ Convolution (64,7,7,subsample = (2,2)) $\rightarrow$ BN $\rightarrow$ ReLU $\rightarrow$
MaxPooling ((3,3), (2,2)) $\rightarrow$ C (3,[16,16,32], BN) $\rightarrow$ I
(3,[16,16,32], BN) $\rightarrow$ I (3,[16,16,32], BN) C (3,[32,32,64], BN)
$\rightarrow$ I (3,[32,32,64], BN) $\rightarrow$ I (3,[32,32,64], BN)
$\rightarrow$ AveragePooling ((7,7)) $\rightarrow$ Dense (10) $\rightarrow$
SoftMax

We used the ADADELTA optimization method with a random grid search for
hyper-parameter optimization. All configurations of the networks were run for
60 iterations apart from the overfit configuration which was run for 100 iterations.

\begin{figure}[hbt]
	\begin{subfigure}[b]{200px}
		\centering
		\includegraphics[width=140px]{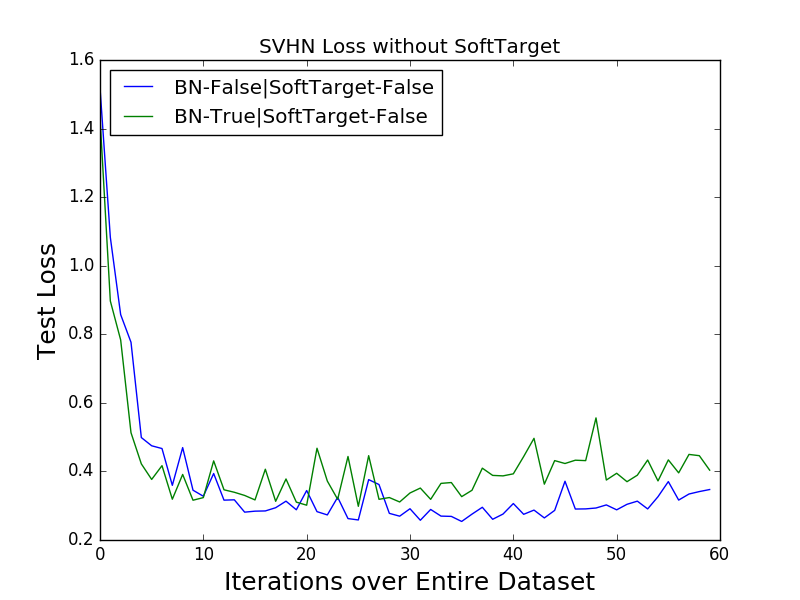}
		\caption{No SoftTarget Regularization }
	\end{subfigure}%
	\begin{subfigure}[b]{200px}
		\centering
		\includegraphics[width=140px]{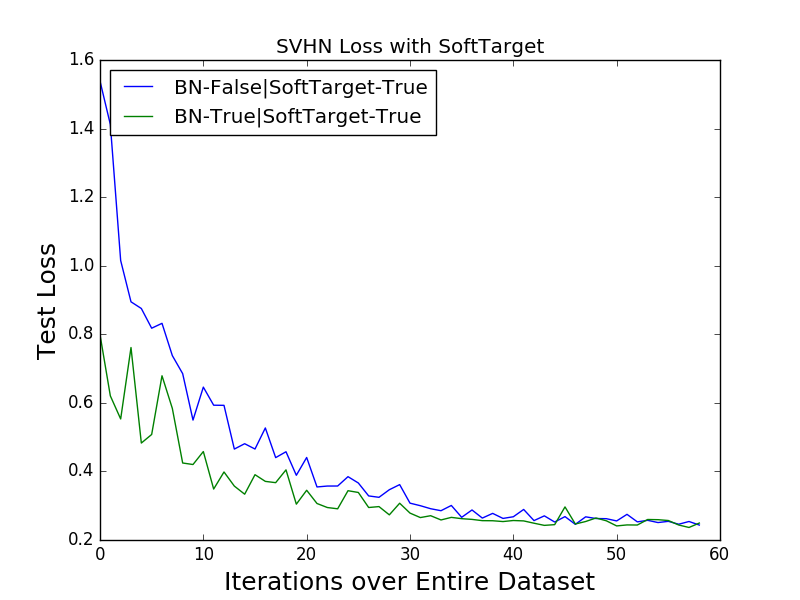}
		\caption{SoftTarget Regularization}
	\end{subfigure}
	\caption{\label{fig:svhn}SoftTarget regularization applied to SVHN dataset.}%
\end{figure}

We reported our results in Table~\ref{tab:svhn} and Figure~\ref{fig:svhn}, as
before giving the minimum test loss and the test loss at the last epoch.
SoftTarget regularized configurations (with and without BN) again scored the
lowest test loss and highest accuracy, compared to Batch Normalization alone.

\begin{table}[hbt]
	\centering
	\caption{\label{tab:svhn}Residual Networks on SVHN}
	\begin{tabular}{ccccc}
		\toprule
		          & No Regularization & BN          & SoftTarget                    & SoftTarget+BN                  \\ \midrule
		Test Loss & 0.254|0.347       & 0.298|0.404 & 0.244|\textcolor{blue}{0.244} & \textcolor{green}{0.237}|0.249 \\ \midrule
		Test Accuracy & 0.929|0.923       & 0.921|0.915 & 0.931|\textcolor{blue}{0.931} & \textcolor{green}{0.932}|0.929 \\ \bottomrule
	\end{tabular}
\end{table}

\subsection{Co-label Similarities}
We claimed that over-fitting begins to occur when co-label similarities that
appeared in the initial stages of training, are not longer present. To test
this hypothesis we compared the covariate matrices of a over-fitted network,
early training stopped networks, and regularized networks. We tested again on
the CIFAR10 dataset, with the same architecture as the previous CIFAR10
experiment, except that the number of filters and dense units were reduced
exactly by two. We compared four configurations: Early (10 epochs), Overfit
(100 epochs), Dropout ($p$=0.2, 100 epochs) and SoftTarget ($n_b = 2,n_t=2,
\beta = 0.7, \gamma = 0.5$, 100 epochs). After training each configuration for
its respected amount we predicted the labels of the training set. We then
calculated a covariance matrix scaled to a range of $[0,1]$ since we are only
interested in the relative co-label similarities. We set the diagonal to all
zeros, as to make it easier to see other relations. The covariance function
used is defined below.

\begin{center}
	\begin{align}
	c_{i,i}   & = 0      \\
	c_{x,y}   & = \frac{\sum_{i=1}^{N}(x_{i}-\bar{x})(y_{i}-\bar{y})}{N-1}      \\
	covs(x,y) & = \frac{c_{x,y} - \min(c_{x,y})}{\max(c_{x,y}) - \min(c_{x,y})}
\end{align}
\end{center}

\begin{figure}[t]

	\begin{center}
		\begin{subfigure}[b]{195px}
			\centering
			\includegraphics[width=100px]{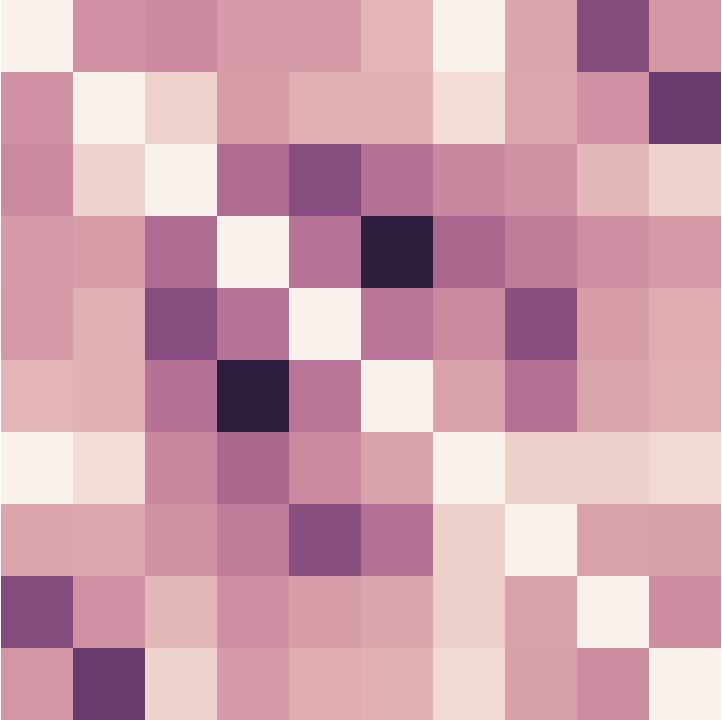}
			\caption{Early Stop}
		\end{subfigure}%
		\begin{subfigure}[b]{195px}
			\centering
			\includegraphics[width=100px]{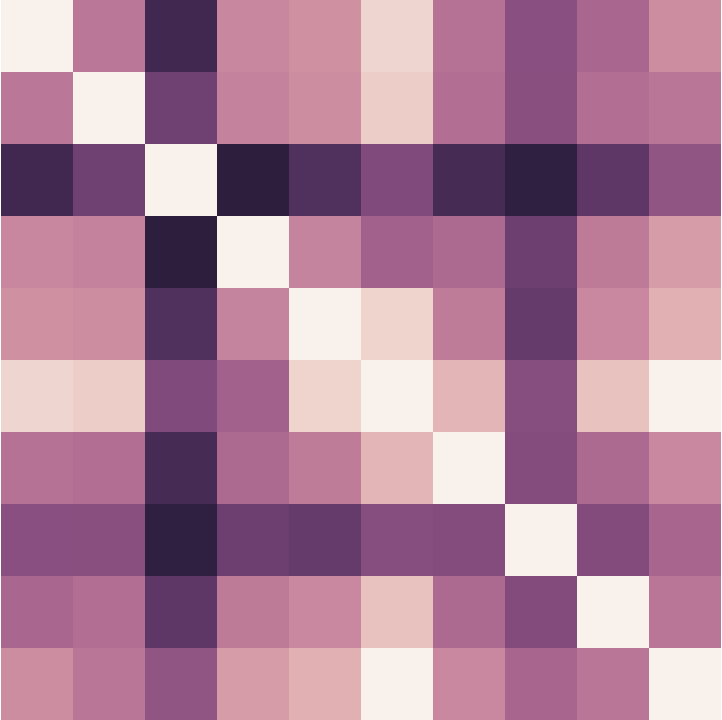}
			\caption{Overfit}
		\end{subfigure}\\
		\begin{subfigure}[b]{195px}
			\centering
			\includegraphics[width=100px]{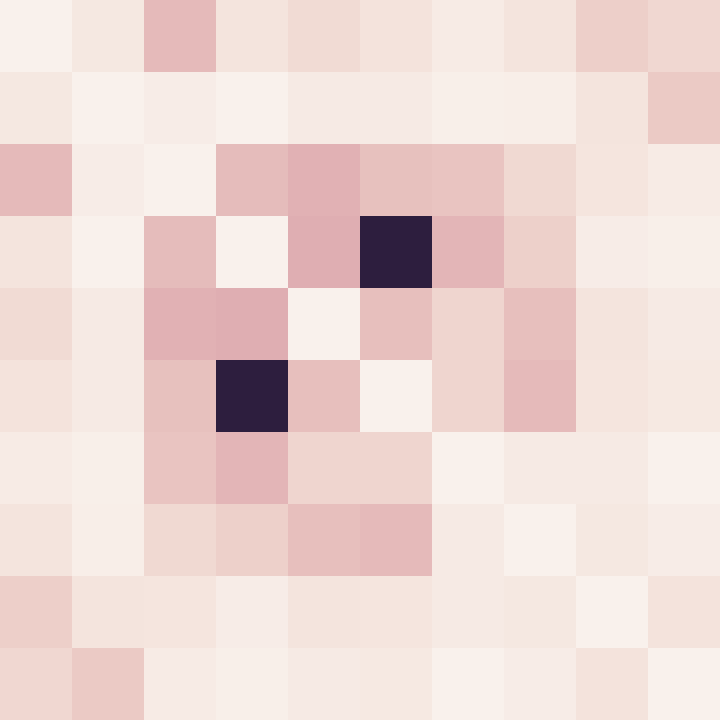}
			\caption{Dropout}
		\end{subfigure}
		\begin{subfigure}[b]{195px}
			\centering
			\includegraphics[width=100px]{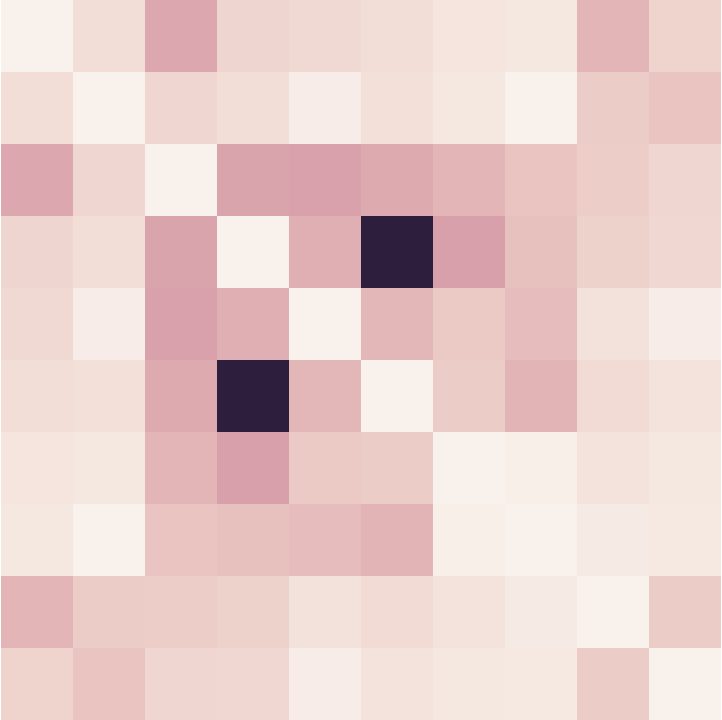}
			\caption{SoftTarget}
		\end{subfigure}
	\end{center}
	\caption{\label{fig:cov}Covariance matrices for the CIFAR10 dataset.}
\end{figure}

We plotted the covariance matrices in Figure~\ref{fig:cov}. For the early
stop case, there we observed the highest covariance between labels 3 and 5,
which correspond to cats and dogs respectively. This intuitively makes sense,
during earlier steps of training, the network learns to first detect
differences between varying entities, such as frog and airplane, and then later
learns to detect subtle difference. It is interesting to note, that this is the
core principle behind prototype theory in human psychology
\citep{Osherson1981}, \citep{Duch1996}, \citep{Rosch1978}. Some concepts are in
nature closer to each other than others. Dog and cat are closer in relation
than frog and airplane, and our regularization method mimics this phenomena.
Another interesting thing to note is that the dropout method of regularization
produces a covariance matrix that is very similar to that produced by
SoftTarget regularization. The phenomena of co-label similarities being
propagated throughout learning is not specific to just SoftTarget
regularization, but regularization in general. Therefore co-label
similarities can be seen as a measure of over-fitting.

\section{Conclusion and Future Work}
In conclusion, we presented a new regularization method based on the
observation that co-label similarities apparent in the beginning of training,
disappear once a network begins to over-fit. SoftTarget regularization reduced
over-fitting as well as Dropout without adding complexity to the network,
therefore reducing computational time, and we provided novel insights into the
problem of over-fitting.

Future work will focus on methods to reduce the number of hyper-parameters
introduced by SoftTarget regularization, as well as providing a formal
mathematical framework to understand the phenomenon of co-label similarities.

\bibliographystyle{iclr2017_conference}
\bibliography{iclr2017_conference}

\begin{thebibliography}{22}
\providecommand{\natexlab}[1]{#1}
\providecommand{\url}[1]{\texttt{#1}}
\expandafter\ifx\csname urlstyle\endcsname\relax
  \providecommand{\doi}[1]{doi: #1}\else
  \providecommand{\doi}{doi: \begingroup \urlstyle{rm}\Url}\fi

\bibitem[Bergstra \& Bengio(2012)Bergstra and
  Bengio]{Bergstra:2012:RSH:2503308.2188395}
James Bergstra and Yoshua Bengio.
\newblock Random search for hyper-parameter optimization.
\newblock \emph{J. Mach. Learn. Res.}, 13\penalty0 (1):\penalty0 281--305,
  February 2012.
\newblock ISSN 1532-4435.
\newblock URL \url{http://dl.acm.org/citation.cfm?id=2503308.2188395}.

\bibitem[Chollet(2015)]{chollet2015}
Fran{\c{c}}ois Chollet.
\newblock {Keras Deep Learning Library}, 2015.
\newblock URL \url{https://github.com/fchollet/keras}.

\bibitem[Duch(1996)]{Duch1996}
W.~Duch.
\newblock {Categorization, prototype theory and neural dynamics}.
\newblock In T.~Yamakawa and G.~Matsumoto (eds.), \emph{Proceedings of the 4th
  International Conference on Soft Computing}, volume~96, pp.\  482--485, 1996.

\bibitem[Grandvalet \& Bengio(2005)Grandvalet and Bengio]{Grandvalet2005}
Yves Grandvalet and Yoshua Bengio.
\newblock {Semi-supervised Learning by Entropy Minimization}.
\newblock \emph{Network}, 17\penalty0 (5):\penalty0 529--536, 2005.

\bibitem[He et~al.(2015)He, Zhang, Ren, and Sun]{He2015}
Kaiming He, Xiangyu Zhang, Shaoqing Ren, and Jian Sun.
\newblock {Deep Residual Learning for Image Recognition}.
\newblock \emph{arXiv}, pp.\  1--12, December 2015.
\newblock URL \url{http://arxiv.org/abs/1512.03385}.

\bibitem[Hinton et~al.(2015)Hinton, Vinyals, and Dean]{Hinton2015}
Geoffrey Hinton, Oriol Vinyals, and Jeff Dean.
\newblock {Distilling the Knowledge in a Neural Network}.
\newblock \emph{arXiv}, pp.\  1--9, 2015.
\newblock URL \url{http://arxiv.org/abs/1503.02531}.

\bibitem[Hunter(2007)]{Hunter2007}
John~D Hunter.
\newblock {Matplotlib: A 2D Graphics Environment}.
\newblock \emph{Computing in Science and Engineering}, 9\penalty0 (3):\penalty0
  90--95, May 2007.

\bibitem[Ioffe \& Szegedy(2015)Ioffe and Szegedy]{Ioffe2015}
Sergey Ioffe and Christian Szegedy.
\newblock {Batch Normalization: Accelerating Deep Network Training by Reducing
  Internal Covariate Shift}.
\newblock \emph{arXiv}, pp.\  1--11, February 2015.
\newblock URL \url{http://arxiv.org/abs/1502.03167}.

\bibitem[Krizhevsky \& Hinton(2009)Krizhevsky and Hinton]{Krizhevsky2009}
Alex Krizhevsky and Geoffrey Hinton.
\newblock {Learning Multiple Layers of Features from Tiny Images}.
\newblock Technical report, University of Toronto, Toronto, ON, 2009.

\bibitem[Krizhevsky et~al.(2012)Krizhevsky, Sutskever, and
  Hinton]{Krizhevsky2012}
Alex Krizhevsky, Ilya Sutskever, and Geoffrey~E Hinton.
\newblock {ImageNet Classification with Deep Convolutional Neural Networks}.
\newblock \emph{Advances In Neural Information Processing Systems}, pp.\
  1097--1105, 2012.

\bibitem[Krogh \& Hertz(1992)Krogh and Hertz]{Krogh1992}
A.~Krogh and J.~a. Hertz.
\newblock {A Simple Weight Decay Can Improve Generalization}.
\newblock \emph{Advances in Neural Information Processing Systems}, 4:\penalty0
  950--957, 1992.

\bibitem[LeCun et~al.(1998)LeCun, Cortes, and Burges]{LeCun1998}
Yann LeCun, Corinna Cortes, and Christopher J~C Burges.
\newblock {The MNIST Database}, 1998.
\newblock URL \url{http://yann.lecun.com/exdb/mnist/}.

\bibitem[Lee(2013)]{Lee2013}
Dong-Hyun Lee.
\newblock {Pseudo-label: The simple and efficient semi-supervised learning
  method for deep neural networks}.
\newblock In \emph{ICML 2013 Workshop: Challenges in Representation Learning
  (WREPL)}, 2013.

\bibitem[Netzer et~al.(2011)Netzer, Wang, Coates, Bissacco, Wu, and
  Ng]{Netzer2011}
Yuval Netzer, Tao Wang, Adam Coates, Alessandro Bissacco, Bo~Wu, and Andrew~Y
  Ng.
\newblock {Reading Digits in Natural Images with Unsupervised Feature
  Learning}.
\newblock In \emph{NIPS Workshop on Deep Learning and Unsupervised Feature
  Learning}, pp.\  1--9, 2011.

\bibitem[Osherson \& Smith(1981)Osherson and Smith]{Osherson1981}
Daniel~N. Osherson and Edward~E. Smith.
\newblock {On the adequacy of prototype theory as a theory of concepts}.
\newblock \emph{Cognition}, 9\penalty0 (1):\penalty0 35--58, January 1981.

\bibitem[Reed et~al.(2014)Reed, Lee, Anguelov, Szegedy, Erhan, and
  Rabinovich]{Reed2014}
Scott Reed, Honglak Lee, Dragomir Anguelov, Christian Szegedy, Dumitru Erhan,
  and Andrew Rabinovich.
\newblock {Training Deep Neural Networks on Noisy Labels with Bootstrapping}.
\newblock \emph{arXiv}, pp.\  1--11, December 2014.
\newblock URL \url{http://arxiv.org/abs/1412.6596}.

\bibitem[Rosch(1978)]{Rosch1978}
Eleanor Rosch.
\newblock {Principles of Categorization}.
\newblock In Eleanor Rosch and Barbara~L. Lloyd (eds.), \emph{Cognition and
  categorization}, pp.\  27--48. Lawrence Erlbaum, Hillsdale, NJ, 1st edition,
  1978.

\bibitem[Scherer et~al.(2010)Scherer, M{\"{u}}ller, and Behnke]{Scherer2010}
Dominik Scherer, Andreas M{\"{u}}ller, and Sven Behnke.
\newblock \emph{{Evaluation of Pooling Operations in Convolutional
  Architectures for Object Recognition}}, pp.\  92--101.
\newblock Springer Berlin Heidelberg, Berlin, Heidelberg, 2010.

\bibitem[Srivastava et~al.(2014)Srivastava, Hinton, Krizhevsky, Sutskever, and
  Salakhutdinov]{Srivastava2014}
Nitish Srivastava, Geoffrey Hinton, Alex Krizhevsky, Ilya Sutskever, and Ruslan
  Salakhutdinov.
\newblock {Dropout: A Simple Way to Prevent Neural Networks from Overfitting}.
\newblock \emph{Journal of Machine Learning Research}, 15:\penalty0 1929--1958,
  2014.

\bibitem[{The Theano Development Team}(2016)]{TheTheanoDevelopmentTeam2016}
{The Theano Development Team}.
\newblock {Theano: A Python framework for fast computation of mathematical
  expressions}.
\newblock \emph{arXiv}, pp.\ ~19, May 2016.
\newblock URL \url{http://arxiv.org/abs/1605.02688}.

\bibitem[Wan et~al.(2013)Wan, Zeiler, Zhang, LeCun, and Fergus]{Wan2013}
Li~Wan, Matthew Zeiler, Sixin Zhang, Yann LeCun, and Rob Fergus.
\newblock {Regularization of Neural Networks using DropConnect}.
\newblock In \emph{Proceedings of the 30th International Conference on Machine
  Learning}, pp.\  109--111, 2013.

\bibitem[Zeiler(2012)]{Zeiler2012}
Matthew~D. Zeiler.
\newblock {ADADELTA: An Adaptive Learning Rate Method}.
\newblock \emph{arXiv}, pp.\  1--6, December 2012.
\newblock URL \url{http://arxiv.org/abs/1212.5701}.

\end{thebibliography}

\end{document}